\documentclass[letterpaper]{article} % DO NOT CHANGE THIS
\usepackage[]{aaai25}  % DO NOT CHANGE THIS
\usepackage{times}  % DO NOT CHANGE THIS
\usepackage{helvet}  % DO NOT CHANGE THIS
\usepackage{courier}  % DO NOT CHANGE THIS
\usepackage[hyphens]{url}  % DO NOT CHANGE THIS
\usepackage{graphicx} % DO NOT CHANGE THIS
\usepackage{natbib}  % DO NOT CHANGE THIS AND DO NOT ADD ANY OPTIONS TO IT
\usepackage{caption} % DO NOT CHANGE THIS AND DO NOT ADD ANY OPTIONS TO IT
\usepackage{algorithm}
\usepackage{algorithmic}

\usepackage{newfloat}
\usepackage{listings}
\usepackage{booktabs}
\usepackage{times}
\usepackage{soul}
\usepackage{amssymb}
\usepackage{graphicx}
\usepackage{color}

\usepackage{multirow}
\usepackage{algorithm}
\usepackage{algorithmic}
\usepackage[misc]{ifsym}

\usepackage{amsmath,amssymb}
\usepackage{algorithm,algorithmic}
\usepackage{textcomp}
\usepackage{graphbox}

\DeclareCaptionStyle{ruled}{labelfont=normalfont,labelsep=colon,strut=off} % DO NOT CHANGE THIS
\floatstyle{ruled}
\newfloat{listing}{tb}{lst}{}
\floatname{listing}{Listing}
\nocopyright 
\title{M2OST: Many-to-one Regression for Predicting Spatial Transcriptomics from Digital Pathology Images}
\author{
    Hongyi Wang\textsuperscript{\rm 1},
    Xiuju Du\textsuperscript{\rm 2},
    Jing Liu\textsuperscript{\rm 2},
    Shuyi Ouyang\textsuperscript{\rm 1},
    Yen-Wei Chen\textsuperscript{\rm 3\footnotemark[1]},
    Lanfen Lin\textsuperscript{\rm 1\thanks{Corresponding authors.}}
}
\affiliations{
    \textsuperscript{\rm 1}Zhejiang University, Hangzhou, China\\
\textsuperscript{\rm 2}Zhejiang Lab, Hangzhou, China\\
\textsuperscript{\rm 3}Ritsumeikan University, Osaka, Japan\\
    chen@is.ritsumei.ac.jp,
    llf@zju.edu.cn
}

\begin{document}

\maketitle

\begin{abstract}
The advancement of Spatial Transcriptomics (ST) has facilitated the spatially-aware profiling of gene expressions based on histopathology images. Although ST data offers valuable insights into the micro-environment of tumors, its acquisition cost remains expensive. Therefore, directly predicting the ST expressions from digital pathology images is desired. Current methods usually adopt existing regression backbones along with patch-sampling for this task, which ignores the inherent multi-scale information embedded in the pyramidal data structure of digital pathology images, and wastes the inter-spot visual information crucial for accurate gene expression prediction. To address these limitations, we propose M2OST, a many-to-one regression Transformer that can accommodate the hierarchical structure of the pathology images via a decoupled multi-scale feature extractor. Unlike traditional models that are trained with one-to-one image-label pairs, M2OST uses multiple images from different levels of the digital pathology image to jointly predict the gene expressions in their common corresponding spot. Built upon our many-to-one scheme, M2OST can be easily scaled to fit different numbers of inputs, and its network structure inherently incorporates nearby inter-spot features, enhancing regression performance. We have tested M2OST on three public ST datasets and the experimental results show that M2OST can achieve state-of-the-art performance with fewer parameters and floating-point operations (FLOPs). 

\begin{links}
\link{Code}{https://github.com/Dootmaan/M2OST}
% \link{Datasets}{https://aaai.org/example/datasets}
% \link{Extended version}{https://arxiv.org/pdf/2409.15092}
\end{links}
\end{abstract}

\section{Introduction}

Digital pathology images, as a kind of Whole Slide Images (WSIs), have witnessed widespread utilization in research nowadays, as they can be more easily stored and analyzed compared to traditional glass slides \cite{niazi2019digital}. However, besides the spatial organization of cells presented in these pathology images, the spatial variance of gene expressions is also very important for unraveling the intricate transcriptional architecture of multi-cellular organisms \cite{rao2021exploring,tian2023expanding,cang2023screening}. As the extended technologies of single-cell RNA sequencing \cite{singlecellsequencing,mrabah2023toward}, ST technologies have been developed recently, facilitating such spatially-aware profiling of gene expressions within tissues \cite{rodriques2019slide,lee2021xyzeq,bressan2023dawn}.

\begin{figure}[t]
\includegraphics[width=0.48\textwidth]{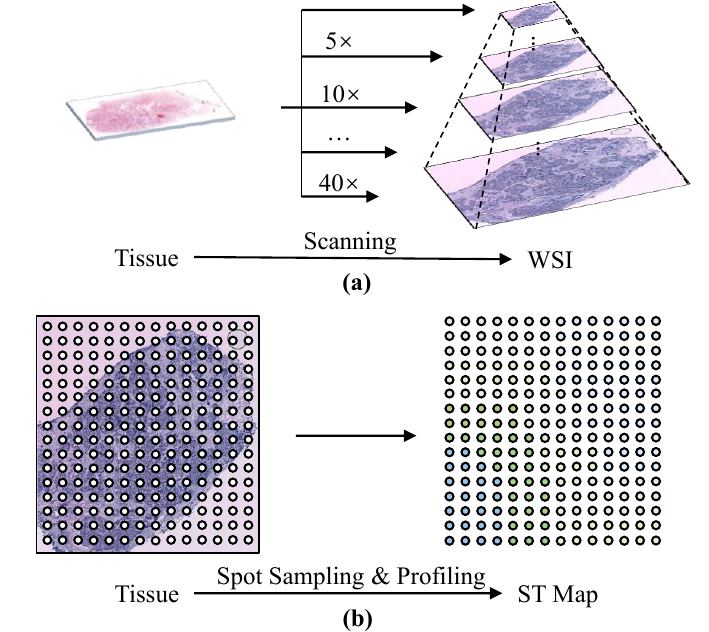}
\caption{(a) WSIs are obtained by scanning the glass slide tissues at different magnifications, resulting in a multi-scale pyramid data structure. (b) ST maps are generated by sampling spots on the glass slide tissues, followed by comprehensive profiling of gene expressions within each sampled spot. } 
% \vspace{-3mm}
\label{fig_wsi_st}
\end{figure}
% \begin{figure}[t]
% \includegraphics[width=\textwidth]{m2o_vs_o2o.pdf}
% \caption{(a) One-to-one regression models optimized with single-level image-label pairs. (b) One-to-one regression models optimized with multi-level image-label pairs. (c) Our proposed many-to-one regression model optimized with multi-level imageset-label pairs. } 
% % \vspace{-3mm}
% \label{m2o_vs_o2o}
% \end{figure}

A detailed illustration of the acquisition process of WSIs and ST maps is presented in Figure~\ref{fig_wsi_st}. As shown, WSIs are obtained by scanning the glass slide tissues at various magnification factors, resulting in a multi-scale hierarchical data structure \cite{ryu2023ocelot}. Correspondingly, ST maps are obtained by firstly sampling spots with a fixed interval on the glass slide tissues. Each spot contains two to dozens of cells depending on different ST technologies \cite{song2021dstg}. Subsequently, the accumulated gene expressions of the cells within the spots are profiled, forming a spatial gene expression map. Such gene expression maps can be used along with their corresponding WSIs for multi-modal computational pathology analysis, leading to higher performance in tasks such as cancer sub-typing and prognosis prediction \cite{hoang2022prediction}. However, despite their rapid evolution, ST technologies have yet to find widespread application in pathological analysis, primarily due to the expensive costs \cite{histogene}. In contrast, WSIs are more economical and accessible as they are routinely generated in clinics \cite{histogene}. Consequently, there is a growing imperative to directly generate ST maps from WSIs at a low cost through deep learning methods \cite{levy2020spatial,weitz2021investigation}. 

Current approaches typically treat the ST prediction problem as a conventional regression problem \cite{stnet,deepspace}, where the network is fed with a WSI patch as input and produces the cumulative gene expression intensities of the cells within the corresponding patch area. In this paradigm, the methods are trained with single-level image-label pairs just like standard regression tasks. This makes them only able to model the relationship between the gene expressions and the images of the maximum magnification, wasting the multi-scale information inherent in WSIs. From a bionic perspective, pathologists often zoom in and zoom out frequently when analyzing WSIs, as each level of WSIs encapsulates distinct morphological information that can be useful for ST predictions \cite{hipt,yarlagadda2023discrete}. For instance, cell-level images can facilitate the evaluation of gene expressions based on cell types, while higher-level images can offer regional morphologies that help determine overall gene intensities. Hence, we propose to conceptualize the ST prediction as a many-to-one modeling problem, in which case multiple images from different levels of WSI are leveraged to jointly predict the gene expressions within the spots. As we notice that the absolute field of view of the microscope will not change during the zooming operations performed by pathologists, we also biomimetically employ a fixed patch size for the pathology patches from different WSI levels in our regression model. In this case, higher-level image patches naturally have a larger receptive field, and thus is able to include more supporting features around the ST spot, compensating for the destroyed cell features on the patch edges during the patch cropping procedure \cite{triplex}. 

Many-to-one-based modeling aims to learn a mapping function from a variable number of inputs to one single output. These multiple inputs can have different shapes or lack semantic alignment, with the goal being to find their common mapping target. It can be used for many tasks, such as multi-phase radiology image analysis \cite{hu2023sam} and label assignment problem \cite{wei2023guide}. Our many-to-one scheme differs from conventional multi-scale methods by offering a structure that can easily scale to accommodate different numbers and different shapes of inputs. For instance, while we primarily present our model in a three-to-one structure, it can be easily adjusted to two-to-one or four-to-one scenarios by removing or adding streams in the pipeline, making it suitable for different WSI scanning technologies. Additionally, during training, model parameters can be partially updated when some levels of inputs are missing, as the model parameters are highly decoupled across the multiple inputs.

Based on this idea, we propose M2OST, a many-to-one-based regression Transformer designed to leverage pathology images at various levels to jointly predict the gene expressions. By incorporating the inter-spot visual information and the multi-scale features within the WSIs, M2OST exhibits the capability to generate more accurate ST maps. Moreover, to optimize the computational efficiency, we further introduce Intra-Level Token Mixing Module (ITMM), Cross-Level Token Mixing Module (CTMM), and Cross-Level Channel Mixing Module (CCMM) to decouple the many-to-one multi-scale feature extraction process into intra-scale representation learning and cross-scale feature interaction processes, which greatly reduces the computational cost without compromising model performance. In summary, our contributions are: 

\begin{enumerate}
    \item We propose to conceptualize the ST prediction problem as a many-to-one modeling problem, leveraging the multi-scale information and inter-spot features embedded in the hierarchically structured WSIs for joint prediction of the ST maps.

    \item We propose M2OST, a flexible regression Transformer crafted to model many-to-one relationships for ST prediction. Its unique design makes M2OST suitable for different many-to-one scenarios, and is robust to input sets with various sequence lengths.

    \item In M2OST, we propose to decouple the multi-scale feature extraction process into intra-scale feature extraction and cross-scale feature extraction, which significantly improves the computational efficiency without compromising model performance.

    \item We have conducted thorough experiments on the proposed M2OST method, and have proved its effectiveness with three public ST datasets.
    
\end{enumerate}

\begin{figure*}[t]
\includegraphics[width=\textwidth]{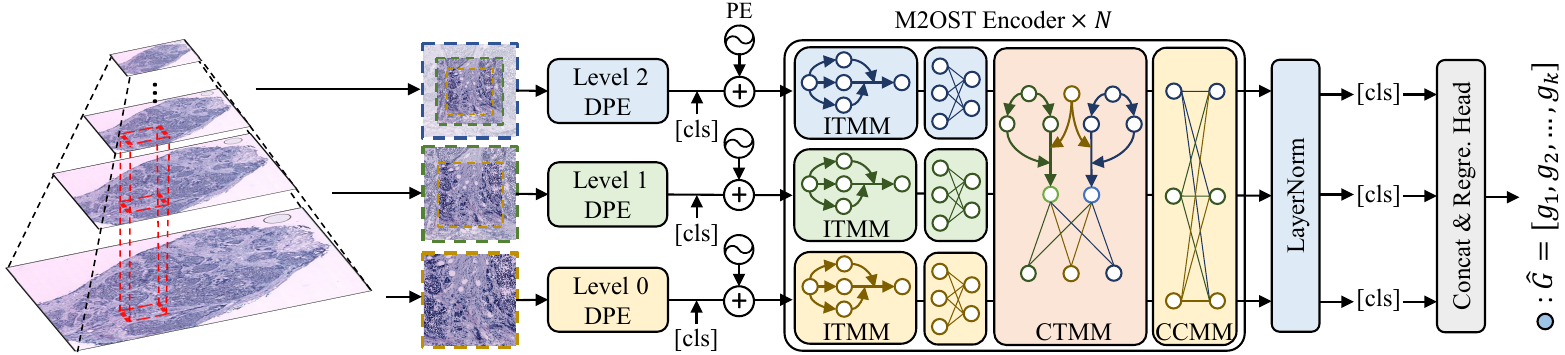}
\caption{A schematic view of the proposed M2OST. Three patch sequences from different WSI levels are fed into the model to jointly predict the gene expressions in the corresponding spot. PE denotes the fully learnable positional embedding in the figure.} 
% \vspace{-3mm}
\label{M2OST_structure}
\end{figure*}

\begin{figure}[t]
\center
\includegraphics[width=0.47\textwidth]{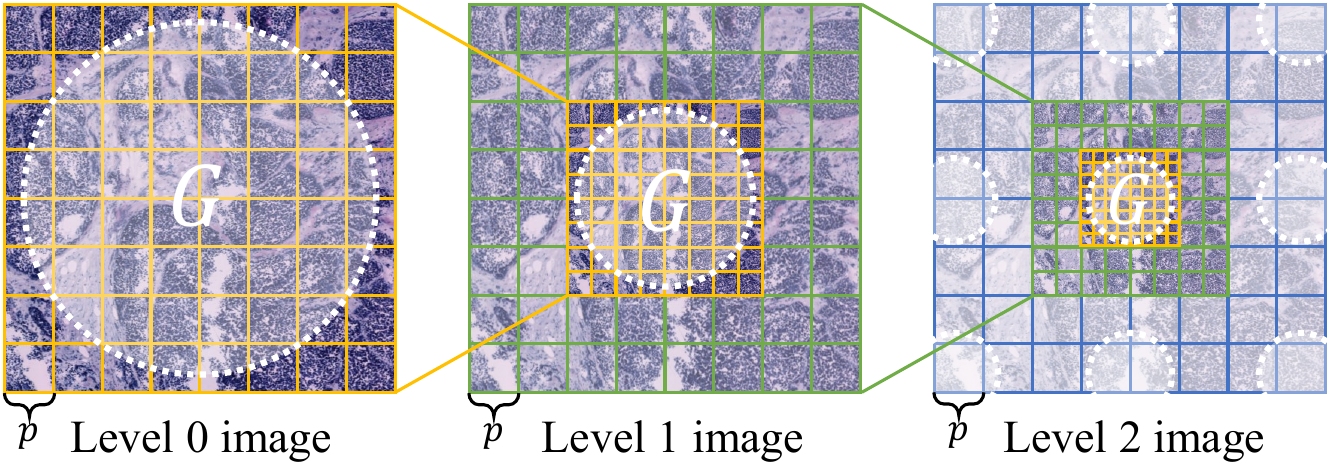}
\caption{DPE used in M2OST. The circle area of $G$ indicates the target ST spot. } 
% \vspace{-3mm}
\label{dpe}
\end{figure}

\section{Related Works}
% \subsection{ST Prediction Based on WSIs}
The prediction of ST maps from WSIs has garnered sustained attention since the inception of ST technologies. ST-Net \cite{stnet} is the first work that attempts to tackle this problem. ST-Net employs a convolutional neural network (CNN) with dense residual connections \cite{resnet,densenet} to predict patch-wise gene expressions. By sequentially processing the patches in a WSI, ST-Net can eventually generate a complete ST map. Similarly, DeepSpaCE \cite{deepspace} adopts a VGG-16 \cite{vgg} based CNN for such patch-level ST prediction, and it introduces semi-supervised learning techniques to augment the training sample pool. More recently, BLEEP \cite{bleep} introduced a contrastive learning approach to align WSI patch features with ST spot embeddings, using K-Nearest Neighbors during the inference stage to mitigate the batch effect in biomedical datasets.

Although these classic CNN backbones have demonstrated considerable success in various vision tasks, their performance has been eclipsed by the advancements achieved with Transformer-based models \cite{ding2023pathology}. HisToGene \cite{histogene} was the first method proposed to leverage vision Transformers \cite{vit} for predicting ST maps. Diverging from the approach of ST-Net and DeepSpaCE, which predict one spot at a time, HisToGene proposes to predict the entire ST map at a time. HisToGene takes the sequenced patches in a WSI as network input and employs the Self-Attention mechanism \cite{transformer} to model the inter-correlations between these patches. Despite the efficiency gained from this slide-level scheme, the performance of HisToGene is constrained by the use of a relatively small ViT backbone, driven by computational limitations. 

Following the path of HisToGene, Hist2ST \cite{hist2st} was then proposed. Combining CNNs, Transformers, and Graph Neural Networks \cite{hamilton2017inductive}, Hist2ST strives to capture more intricate long-range dependencies. Like HisToGene, Hist2ST is also a slide-level method that uses the patch sequence as input to directly generate the gene expressions of all spots in an ST map. However, the complexity of its model structure results in considerable FLOPs and model size, elevating the risk of over-fitting. 

Contrary to the prevalent belief in the necessity of inter-spot correlations for predicting ST maps, iStar\cite{istar} argues that gene expressions within a spot are logically related only to its corresponding patch area, thus reverting to a spot-level training scheme. It adopts HIPT\cite{hipt}, a hierarchical Vision Transformer pre-trained on large-scale WSI datasets for non-trainable slide-level feature extraction, and utilizes a simple MLP to fit the mapping relation from the feature maps to the ST spots, achieving state-of-the-art performance. However, as the feature extraction stage of iStar is unlearnable, it still leaves space for performance improvements. Building on this insight, in our proposed M2OST, we also adhere to the patch-level scheme, predicting a single spot at a time to ensure the independence and accuracy of each prediction.

\section{Methodology}

\subsection{Problem Formulation}
In M2OST, we use $I_0$, $I_1$, and $I_2 \in R^{3\times H\times W}$ to represent the three input images of different levels, where $I_i$ denotes the pathology image patch from level $i$, and $H$, $W$ represents the image height and width, respectively. The observed gene expressions in each spot are denoted as $G=\{g_1, g_2, ..., g_k\}$, where $k$ is the total number of genes. The goal is to minimize the mean squared error (MSE) between $\hat{G}=\mathrm{M2OST}(\{I_0,I_1,I_2\}|\theta_0,\theta_1,\theta_2)$ and $G$ by optimizing the network parameters $\theta_0$,$\theta_1$, and $\theta_2$ of each stream.

\begin{figure}[t]
\center
\includegraphics[width=0.47\textwidth]{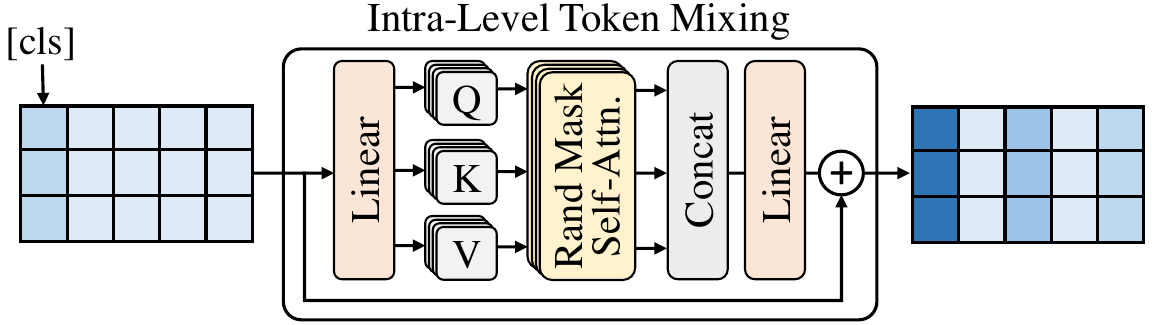}
\caption{The network structure of ITMM. This module needs to be applied to each level's sequence separately.} 
% \vspace{-3mm}
\label{Intra_level_mixing}
\end{figure}

\begin{figure*}[t]
\center
\includegraphics[width=\textwidth]{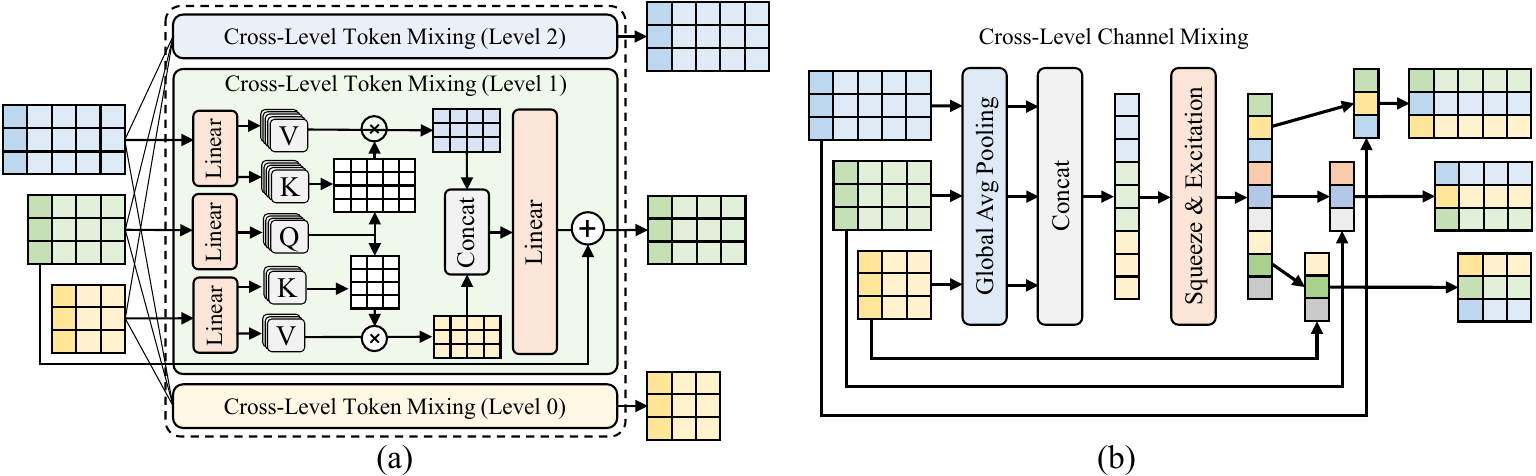}
\caption{(a) The network structure of CTMM. (b) The network structure of CCMM.} 
% \vspace{-3mm}
\label{ctmm_ccmm}
\end{figure*}

\subsection{Overview of M2OST}
A schematic view of the proposed M2OST is presented in Figure~\ref{M2OST_structure}. Upon receiving the multi-scale pathology image patches from three different levels, M2OST initially sends them into our proposed Deformable Patch Embedding (DPE) layers to realize adaptive token generation. After appending $[cls]$ token to each sequence, intra-scale representation learning within each sequence is first performed using ITMM. Then, CTMM is introduced to facilitate cross-scale information exchange between the different inputs, followed by CCMM mixing the channels in a squeeze-and-excitation way. This multi-scale feature extraction module is termed the M2OST Encoder and is iterated $N$ times within M2OST. Finally, the three $[cls]$ tokens are concatenated to be fed into the linear regression head for the ST spot prediction.

\subsection{Deformable Patch Embedding}

Although high-level pathology patches provide more nearby visual information of the target ST spot, the central image area that directly maps to the target spot should still be primarily focused on. To emphasize the in-spot features during many-to-one modeling, we introduce DPE to generate fine-grained in-spot tokens and coarse-grained surrounding tokens. As shown in Fig~\ref{dpe}, apart from using patch size $p$ on $I_0$, $I_1$, and $I_2$ to generate the basic tokens in a weight-sharing manner, DPE also adopts $\frac{p}{2}$ and $\frac{p}{4}$ patch sizes on the higher-level pathology images to ensure the central area of the patches receives the most attention. Eventually, DPE converts the three input images into $L\times C$, $2L\times C$, and $3L\times C$ sequences $S_0,S_1,S_2$, where $L=\frac{HW}{p^2}$ represents the sequence length and $C$ denotes the number of channels after embedding.

\subsection{Intra-Level Token Mixing}
After patch embedding, a $[cls]$ token is appended to the beginning of each sequence. Then, a fully learnable positional embedding is added to each sequence to encode the positional information for the multi-scale tokens generated by DPE. After that, each level's sequence data undergoes processing through ITMM to extract the intra-level features, whose structure is mainly based on ViT \cite{vit}, as depicted in Figure~\ref{Intra_level_mixing}. ITMM uses the Random Mask Self-Attention trick \cite{dropattention} to enhance its generalization ability.

\subsection{Cross-Level Token Mixing}

After the intra-level feature extraction, the acquired representations are amalgamated into CTMM for cross-level information exchange. Given that the sequence lengths of $S_0$, $S_1$, and $S_2$ are different, these data can not be directly fused together for simple information exchange. In the meantime, to most possibly retain the independence of the network parameters $\theta_0$, $\theta_1$, and $\theta_2$, CTMM introduces fully-connected cross-level attention to realize the goal, as illustrated in Fig~\ref{ctmm_ccmm}(a). 

% Specifically, $S_0$, $S_1$, and $S_2$ are firstly sent into linear layers to produce their respective $K$, $Q$, $V$. For different levels' sequence, they use their own $Q$ to compute attention scores with other levels' $K$ and $V$ to introduce cross-scale information into themselves. 

Let $M$ represents the total number of input sequences, and $K_i$, $Q_i$, $V_i$ denotes the hidden representation extracted from $S_i$, CTMM can be mathematically defined as: 
\begin{equation}
    CTMM(S_i)= \sum_{0\le j<M}^{j\neq i} \omega_j^i \cdot \sigma (\frac{Q_iK_j^T}{\sqrt{d_K}})V_j,
\end{equation}
\noindent where $\sigma$ represents the Softmax operation, and $\omega_j^i$ is the learnable weights indicating how much the $j$-th level information contributes to the $i$-th level CTMM output. Finally, $M$ sequences are obtained with their shape unchanged, and subsequently sent into CCMM for further channel mixing.

\subsection{Cross-Level Channel Mixing}

CCMM is used to explicitly facilitate the channel interaction between multi-level sequences. Since the input sequences are still of different lengths, we design a length-insensitive channel mixing method for CCMM to address this issue, which is presented in Fig~\ref{ctmm_ccmm}(b). Inspired by the squeeze-and-excitation operation in \cite{danet}, we first use global average pooling for each sequence to compress their sequential information into one token. Then, we combine these tokens from different levels together and use a squeeze-and-excitation operation to obtain the cross-level channel attention scores. After that, the scores are split and multiplied back to their respective input sequences, leading to channel-level cross-scale information exchange. 

In summary, as afore-presented, every module of M2OST is insensitive to sequence length, and can be easily scaled to handle different numbers of input by removing or adding streams to the pipeline. 

% \subsection{Regression Head}
% Upon the completion of $N\times$ M2OST Encoders, the multi-scale features are properly handled. Next, the regression head is applied on the three $[cls]$ tokens to predict the gene expression of their corresponding common ST spot. The current M2OST model uses one linear layer on the fused $[cls]$ token to project the $1 \times 3C$ feature vector into $k$-channel gene expression vector $\hat{G}$. MSE loss is then computed and back-propagated to optimize the network parameters $\theta$ in an end-to-end manner. 

% Finally, We provide the pseudo-code of M2OST pipeline in Algorithm~\ref{alg:algorithm}, where $PE_i$ denotes the fully-learnable positional embedding for the $i$-th level input sequence.

\begin{table}[t]
\renewcommand\tabcolsep{0.7pt}
\center
\resizebox{1\linewidth}{!}{
\begin{tabular}{@{}ccccccccc@{}}
\toprule
\multirow{2}{*}{DPE} & \multirow{2}{*}{ITMM} & \multirow{2}{*}{CTMM} & \multirow{2}{*}{CCMM} & \multicolumn{3}{c}{PCC (\%)}                     & \multirow{2}{*}{\begin{tabular}[c]{@{}c@{}}Param\#\\ (M)\end{tabular}} & \multirow{2}{*}{\begin{tabular}[c]{@{}c@{}}FLOPs\\ (G)\end{tabular}} \\ \cmidrule(lr){5-7}
  &    &     &      & HBC            & HER2+          & cSCC    &  &    \\ \midrule
  $\checkmark$ &   $\checkmark$      &     $\checkmark$    &   $\checkmark$      & \textbf{48.07} & \textbf{44.17}     & \textbf{50.50}      & \textbf{6.81}    & 2.24 \\
  &     $\checkmark$      &     $\checkmark$    &   $\checkmark$      & 47.13    & 43.10 & 49.35 & 7.76 & \textbf{1.23}   \\
 &      &     $\checkmark$      &     $\checkmark$    & 47.03   & 42.99     & 49.48  & 12.88    & 2.16      \\ 
 & & &   $\checkmark$    & 46.34   & 42.57     & 48.81  & 10.66    & 1.72       \\
 & & &  & 46.12   & 42.55     & 48.66  & 10.66     & 2.07      \\
 \bottomrule
\end{tabular}
}
\caption{Ablation study results based on substituting components of M2OST into others. }
\label{ablation_decouple}
\end{table}

\begin{table}[t]
\renewcommand\tabcolsep{0.5pt}
\center
\resizebox{1\linewidth}{!}{
\begin{tabular}{@{}ccccccccc@{}}
\toprule
\multicolumn{3}{c}{Input} & \multicolumn{2}{c}{HBC} & \multicolumn{2}{c}{HER2+} & \multicolumn{2}{c}{cSCC} \\ \cmidrule(l){1-3}\cmidrule(l){4-5} \cmidrule(l){6-7} \cmidrule(l){8-9} 
Lvl 0   & Lvl 1  & Lvl 2  & PCC(\%)        & RMSE       & PCC(\%)         & RMSE        & PCC(\%)        & RMSE        \\ \midrule
\checkmark  & & & 46.92        &  3.17       &   43.12            &  3.06         &   49.31           &   3.60        \\
 & \checkmark &        &   45.23          &  3.18          &   42.56          & 3.11            &  48.27          &  3.81  \\
 &        & \checkmark &  41.04          &  3.25          &   40.01          & 3.21            &  45.29          &  3.89           \\ \midrule
 \checkmark & \checkmark & &   47.32   &  3.16    &   43.31        &  3.05      &   50.02   &   3.47   \\
\checkmark &  & \checkmark   &    46.94   &  3.17     &   42.98       &  3.10  &   49.73   &   3.61   \\
 & \checkmark & \checkmark &    45.62     &  3.20  &   42.67          & 3.10   &  49.11  &  3.80    \\ \midrule
\checkmark & \checkmark & \checkmark & \textbf{48.07}      & \textbf{3.16}        &   \textbf{44.17}            & \textbf{2.87}          & \textbf{50.50}       &  \textbf{3.45}     \\
        \bottomrule
\end{tabular}
}
\caption{Ablation study on the input combinations of M2OST.}
\label{m2o_ablation}
\end{table}

\section{Experiments}

\subsection{Datasets and Metrics}

In our experiments, we utilized three public datasets to evaluate the performance of the proposed M2OST model.

The first one is the human breast cancer (HBC) dataset \cite{dataset1}. This dataset contains 30,612 spots in 68 WSIs, and each spot has up to 26,949 distinct genes. The spots in this dataset exhibit a diameter of 100 µm, arranged in a grid with a center-to-center distance of 200 µm.

The second dataset is the human HER2-positive breast tumor
dataset \cite{dataset2}. This dataset consists of 36 pathology images and 13,594 spots, and each spot contains 15,045 recorded gene expressions. Similar to the previous dataset, the ST data in this dataset also features a 200µm center-to-center distance between each captured spot with the diameter of each spot also being 100µm. 

The third dataset is the human cutaneous squamous cell carcinoma (cSCC) dataset \cite{dataset3}, which includes 12 WSIs and 8,671 spots. Each spots in this dataset have 16,959 genes profiled. All the spots have a diameter of 110µm and are arranged in a centered rectangular lattice pattern with a center-to-center distance of 150µm.

% \begin{algorithm}[tb]
%     \caption{M2OST for ST prediction}
%     \label{alg:algorithm}
%     \textbf{Input}: Pathology images of three levels $I_0$, $I_1$, $I_2$\\
%     \textbf{Parameter}: the repeated times $N$ of M2OST Encoder, number of head $n\_head$ in ITMM, and the channels $C$ for DPE.\\
%     \textbf{Output}: The gene expressions in the corresponding spot of the images
%     \begin{algorithmic}[1] %[1] enables line numbers
%         \STATE $S_0=Concat(DPE_0(I_0|C),[cls])$.
%         \STATE $S_1=Concat(DPE_1(I_1|C),[cls])$.
%         \STATE $S_2=Concat(DPE_2(I_2|C),[cls])$.
%         \STATE $S_0=S_0+PE_0$.
%         \STATE $S_1=S_1+PE_1$.
%         \STATE $S_2=S_2+PE_2$.
%         \FOR{$i$ in range($N$)}
%         \STATE $S_0, S_1, S_2=Split(Norm(Concat(S_0, S_1, S_2)))$
%         \STATE $S_0=ITMM_0^i(S_0|n\_head)$.
%         \STATE $S_1=ITMM_1^i(S_1|n\_head)$.
%         \STATE $S_2=ITMM_2^i(S_2|n\_head)$.
%         \STATE $S_0, S_1, S_2=Split(Norm(Concat(S_0, S_1, S_2)))$
%         \STATE $S_0, S_1, S_2=CTMM^i(S_0, S_1, S_2)$
%         \ENDFOR
%         \STATE $S_0, S_1, S_2=Split(Norm(Concat(S_0, S_1, S_2)))$
%         \STATE $[cls]=Concat(S_0[0,:],S_1[0,:],S_2[0,:])$
%         \STATE $\hat{G}=Linear([cls])$
%         \STATE \textbf{return} $\hat{G}$
%     \end{algorithmic}
% \end{algorithm}

We employ the mean values of Pearson Correlation Coefficients (PCC) and Root Mean Squared Error (RMSE) of the spots to evaluate the regression accuracy. Mathematically, PCC can be described as:
\begin{equation}
PCC=\frac{Cov (G, \hat{G})}{Var (G) \cdot Var (\hat{G})},
\end{equation}
\noindent where $Cov(\cdot)$ is the covariance, $Var(\cdot)$ is the variance, $G$ is the ground truth gene expressions of a spot, $\hat{G}$ is the corresponding predicted result.
% Subsequently, RMSE can be mathematically defined as:
% \begin{equation}
% RMSE=\sqrt{\frac{1}{k}\sum^k_{i=1}(g_i-\hat{g_i})^2},
% \end{equation}
% \noindent where $G$ is the ground truth gene expressions of a spot, $\hat{G}$ is the corresponding predicted result, $k$ is the total number of the genes in $\hat{G}$, and $\hat{g_i}$ denotes the $i$-th gene in $\hat{G}$.

\subsection{Implementation Details}

Given the inherently sparse nature of the ST map, we filter out less-variable genes in each dataset based on the criteria outlined in \cite{stnet}, eventually preserving 250 spatially variable genes per dataset for training. As for the pre-processing procedures, they are also kept identical to those described in \cite{stnet}. Specifically, we normalize the gene expression counts for each spot by dividing them by the sum of expressions within that spot, then multiplying the result by a scale factor of 1,000,000. The normalized values are subsequently transformed using the natural logarithm, calculated as log(1 + $x$), where $x$ is the normalized count.

For all datasets, we use a patch size of 224$\times$224 (which covers around 110µm$\times$110µm in the pathology image) for each spot on level 0 pathology image, and the patch size $p$ is set to 16 accordingly. In each dataset, 60\% of the WSIs and their corresponding ST maps are used for training, 10\% for validation, and the remaining 30\% for testing. All the methods are trained with Adam \cite{adam} optimizer with a learning rate of 1e-4 for 100 epochs. Batch size is 96 for patch-level methods and 1 for slide-level methods. The hyper-parameters of M2OST are the model width, model depth, and the number of heads in self-attention. The three hyper-parameters were tuned following the goal of surpassing other methods with minimal model size. Specifically, the M2OST Encoder is repeated 4 times (i.e, model depth), the embedding channel is 192 (i.e., model width), and the number of head for the self-attention operation in ITMM is set to be 3. A larger model size can lead to even better ST regression performance but the computational cost will also be higher. All the methods are trained on two Nvidia RTX A6000 (48G) GPUs.

\begin{table*}[t]
\renewcommand\tabcolsep{3pt}
\centering
\begin{tabular}{@{}ccccccccc@{}}
\toprule
\multirow{2}{*}{Methods} & \multicolumn{2}{c}{HBC} & \multicolumn{2}{c}{HER2+} & \multicolumn{2}{c}{cSCC} & \multirow{2}{*}{\begin{tabular}[c]{@{}c@{}}Parameter\\ Count (M)\end{tabular}} & \multirow{2}{*}{\begin{tabular}[c]{@{}c@{}}FLOPs \\(G)\end{tabular}} \\ \cmidrule(lr){2-3} \cmidrule(lr){4-5} \cmidrule(lr){6-7}
                         & PCC(\%)      & RMSE     & PCC(\%)       & RMSE      & PCC(\%)      & RMSE      &                                                                                &                                                                                  \\ \midrule

ResNet50 \cite{resnet}                & 47.10        &  3.17        &   43.33            &  3.04         &   49.34           &   3.60        & 24.02    &  4.11   \\
ViT-B/16 \cite{vit}  & 46.67  &  3.17  & 43.78   &  3.09  &  49.01   & 3.77  & 57.45  &  11.27  \\
Swin-T \cite{swintransformer}  & 44.52  &  3.29 & 37.67  &  3.57  &  48.83   & 3.74  & 19.02  &  2.96  \\
ConvNeXt-T \cite{convnext}  & 47.25  &  3.16  & 43.56   &  3.07 &  \underline{50.08}   & \underline{3.49} & 27.99  &  4.46  \\
CrossViT \cite{crossvit}  & 47.46  &  3.16  & 43.90   &  3.04 &  49.51   & 3.55 & 26.27 &  4.85  \\
\midrule
DeepSpaCE \cite{deepspace}                 & 46.01        & 3.19         &  42.57             &   3.17        & 48.99             & 3.73          & 135.29   & 15.48   \\
ST-Net \cite{stnet}    &  \underline{47.78} & \underline{3.16}         & 43.01  & 3.07          &  49.37   &  3.58 & \underline{7.21}    &   \underline{2.87}   \\
HisToGene \cite{histogene}               &   44.76 & 3.20 & 36.97         &   3.62         &   45.71 &  3.93         &    187.99   &  135.07  \\
Hist2ST \cite{hist2st}  &    45.00  &   3.18  &  40.02  & 3.06      &  46.71   & 3.88     &  675.50  &  1063.23 \\
BLEEP \cite{bleep}  &  47.02  &  3.17 & 43.53 & 3.05     &  49.60  & 3.59  &  24.18  & 4.19 \\
% TRIPLEX \cite{triplex}  & - & -&- & -  &-& -  & 24.31 & - \\
HIPT/iStar \cite{hipt,istar}  & 47.60  &  3.16  & \underline{43.92}  & \underline{3.01}  &  49.73   & 3.52  & 24.59  &  5.13  \\ \midrule
M2OST (Ours)            & \textbf{48.07}      & \textbf{3.16}        &   \textbf{44.17}            & \textbf{2.87}          & \textbf{50.50}       &  \textbf{3.45}       & \textbf{6.81}        & \textbf{2.24}                                      \\ \bottomrule
% M2OST-\underline{B}ase (Ours)            & \underline{-}        & \underline{-}         &  \underline{-}           & \underline{-}          &  \underline{-}  &  \underline{-}         & 20.44    & 6.30   \\ 
% M2OST-\underline{L}arge (Ours)            &   \textbf{-}     &   \textbf{-}       &   \textbf{-}            &  \textbf{-}         &  \textbf{-}            &  \textbf{-}        & 60.28   &  17.56                                                                          \\ \bottomrule
\end{tabular}
\caption{Experimental results of comparing M2OST with other ST or non-ST methods. The best results are marked in \textbf{bold}, and the second-best results are \underline{underlined}. }
\label{comparison_with_other_methods} 
\end{table*}

\subsection{Ablation Study}

\subsubsection{Study on the M2OST Model Structure.}

To verify the effectiveness and efficiency of M2OST, we have conducted a thorough ablation study on its network structure, of which the experimental results are presented in Table~\ref{ablation_decouple}. We begin by replacing DPE with ordinary patch embedding layers, which leads to a notable decrease in PCC of all three datasets, namely 0.94\%, 1.07\%, and 1.15\%. Although the FLOPs dropped due to the reduced input sequence length, the parameter counts increased because of the absence of the weight-sharing mechanism used in DPE. Such experimental results prove the effectiveness of the adaptive patch embedding in DPE.

Then, we substitute the three ITMMs into one unified Self-Attention to directly process the concatenated sequences (the three sequences are of the same length without DPE, so they can be directly concatenated), destroying the decoupled design in M2OST. It is observed that the parameter count dramatically increased, but the model performance did not benefit from it, which validates the efficiency of using ITMM to decouple the multi-scale feature extraction process in M2OST. We further remove CTMM from M2OST, using simple concatenation for cross-level feature fusion. This time, the parameter count did not drop much, while the performance suffered a further decline. This indicates that CTMM is necessary for processing such many-to-one modeling problems, where each sequence may contain different semantic information that cannot be fused by simple concatenation. We have also tried using summation to replace CTMM, but it even fails to outperform the concatenation scheme. 

Finally, we replaced CCMM with ordinary fully connected layers, and the FLOPs increased while the performance did not change much. This illustrates the effectiveness of CCMM in performing channel mixing for sequences of different lengths in M2OST.

\subsubsection{Study on the Input Combinations for M2OST.} Using M2OST as the backbone, various input combinations were fed into the model to verify the effectiveness of our many-to-one design. We kept the network width and depth identical for different combinations of inputs to ensure fairness during comparison, which also leads to similar parameter counts and FLOPs of the compared methods. The experimental results are summarized in Table~\ref{m2o_ablation}. 

Analysis of the table reveals that when employing M2OST as a one-to-one-based method, using level 0 pathology images yields optimal results across all three datasets. This is attributed to the comprehensive high-frequency information present in the level 0 pathology images, validating that the gene expression in a spot is primarily related to its corresponding tissue area. In this case, M2OST also did not surpass other one-to-one-based methods such as ResNet-50 and ST-Net when referring to the results in Table~\ref{comparison_with_other_methods}, which is mainly due to its smaller model size. Nonetheless, after introducing level 1 and level 2 image patches as additional inputs, the PCC of M2OST increases to 48.07\%, 44.17\%, and 50.50\% on the three datasets, achieving state-of-the-art performance. This illustrates the effectiveness of the many-to-one scheme in M2OST, proving that introducing the multi-scale and surround-spot visual information for ST prediction can improve the model accuracy.

\begin{figure*}[t]
\includegraphics[width=\textwidth]{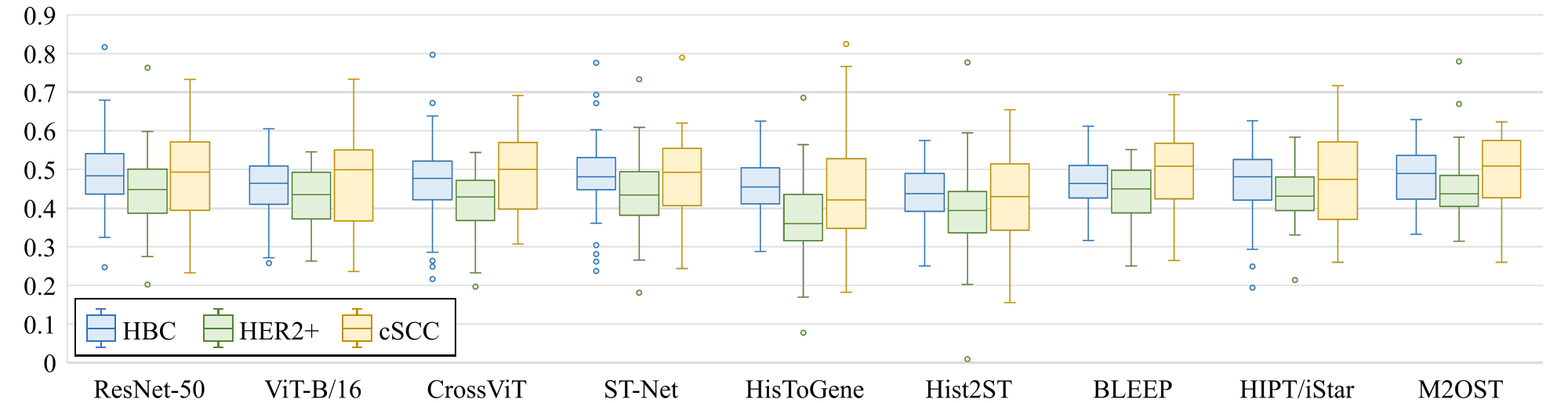}
\caption{The box-plot of different methods' test PCC on the three datasets.} 
% \vspace{-3mm}
\label{boxplot}
\end{figure*}

\begin{figure*}[t]
\includegraphics[width=\textwidth]{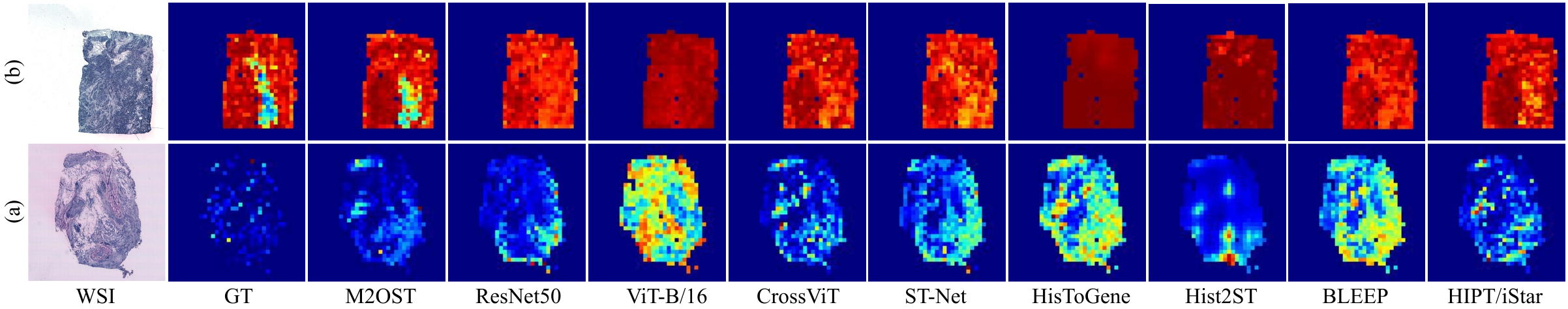}
\caption{(a) Visualization of the ST map after PCA. (b) Visualization of the spatial distribution of the DDX5 gene.} 
% \vspace{-3mm}
\label{visualization_st}
% \vspace{-1.5mm}
\end{figure*}

\subsection{Experimental Results}

\subsubsection{Overview of the Experimental Results.} The experimental results of the comparison between M2OST and other methods are presented in Table~\ref{comparison_with_other_methods}. 
This table provides detailed insights into the PCC and RMSE on various datasets of different methods, along with their parameter count and FLOPs. 
Analysis of the experimental results reveals that M2OST achieves superior performance with fewer FLOPs and a reduced parameter count. In comparison to ST-Net, which features 0.40M more parameters and 0.63G more FLOPs, M2OST surpasses its performance on HER2+ and cSCC datasets by 1.16\% and 1.13\% PCC, respectively.

% It should be noted that the FLOPs of HisToGene and Hist2ST may vary depending on the dataset. This is because HisToGene and Hist2ST are slide-level methods, and their FLOPs are influenced by the length of sequences. Consequently, the table reports the highest GPU memory consumption among all three datasets for these two methods.

% Additionally, a visual comparison between these methods is presented in Figure~\ref{performance_visual_comparison}, where their performance is evaluated based on the mean PCC across the three datasets and the training GPU memory consumption is also considered. The training GPU memory is computed based on a batch size of 96 for patch-level methods and 1 for slide-level methods. Notably, M2OST-S outperforms all other methods in terms of mean PCC and achieves significant savings of GPU computational cost during training, making it the most cost-effective method among all the methods. These results underscore the effectiveness and efficiency of the proposed M2OST.

\subsubsection{Comparison between M2OST and One-to-one Multi-Scale Methods.} 
In Table~\ref{comparison_with_other_methods}, we also have some comparisons with ordinary one-to-one multi-scale methods, such as CrossViT and HIPT/iStar. Compared with the vanilla ViT, CrossViT significant improvement in ST regression performance, confirming the value of incorporating multi-scale information for this task. However, since CrossViT is limited in its ability to fully utilize inter-spot information, it falls short of surpassing the performance of our proposed M2OST model.

In the case of iStar, the model achieved an even higher prediction accuracy for ST, underscoring the effectiveness of HIPT in extracting multi-scale features from WSIs. However, due to HIPT's hierarchical ViT architecture, training the model end-to-end is computationally expensive. As a result, iStar employs frozen HIPT weights to generate WSI features for ST prediction, which might compromise feature extraction performance. Furthermore, our observations (based on the official code release) indicate that iStar requires significantly more processing time during inference. This increased time is primarily attributed to its multi-scale feature extraction process, which operates patch by patch and scale by scale. When we limited M2OST's batch size to match iStar's GPU memory consumption, M2OST demonstrated an inference speed that was 100$\times$ faster than iStar's for ST regression. Despite this remarkable efficiency, M2OST still outperformed iStar, highlighting the superiority of end-to-end training in ST prediction and validating the effectiveness of our model design.

\subsubsection{Comparison between Patch-Level and Slide-Level ST Methods.}
From Table~\ref{comparison_with_other_methods}, it is also observed that the slide-level ST methods fail to outperform patch-level methods on all three datasets. Among the slide-level methods, Hist2ST does surpass HisToGene due to its larger model size, but the extra FLOPs and the dramatic parameter count diminish the significance of this performance improvement. When compared to baseline patch-level methods such as ST-Net, the PCC of Hist2ST is 2.78\%, 2.99\%, and 2.66\% lower on the three datasets respectively. This suggests that the gene expressions of a spot are primarily related to its corresponding tissue area, and introducing inter-spot correlations does little to enhance prediction accuracy.
Nevertheless, slide-level methods still possess the advantage of being more efficient in generating entire ST maps. With a refined network design, they still have the potential of achieving a competitive regression accuracy.

% \begin{figure*}[t]
% \includegraphics[width=\textwidth]{per_gene_visualization.pdf}
% \caption{(a) Visualization of the spatial distribution of the DDX5 gene. (b) Visualization of the spatial distribution of the BRD2 gene.} 
% % \vspace{-3mm}
% \label{per_gene_visualization}
% \end{figure*}

% \subsection{Stability Analysis}

% To assess the regression stability of various methods, we further conducted a comprehensive box-plot comparison, as depicted in Figure~\ref{boxplot}. Comparing the three primary patch-level regression baselines, namely ResNet-50, ViT-B/16, and ST-Net, it is shown that ST-Net showcased superior overall stability in comparison to the other methods despite exhibiting a higher number of outliers. Transitioning to slide-level methods HisToGene and Hist2ST, the stability experienced a notable decline, particularly evident in the cSCC dataset. Lastly, introducing M2OST into the comparison revealed that the proposed M2OST method demonstrated the most stable predictions across all considered methods, validating the effectiveness of our network design.

% \subsection{Study on the Transferring Ability}

% We also tested the transferring ability of different methods to make sure they can adapt to unseen data in real-life clinical use.

\subsubsection{Statistical Significance and Deviation Analysis.}

A paired T-test for M2OST predictions has been conducted to ensure the statistical significance of the experimental results, and it is observed that p-value\textless 0.05 holds for all other methods. We have also presented a boxplot in Fig~\ref{boxplot}, and it is shown that M2OST demonstrated the most stable predictions across all considered methods, validating the effectiveness of its network design.

\subsubsection{Visualization Analysis.}

Finally, we present some visualization results in Figure~\ref{visualization_st} to make an intuitive comparison of the methods. In Figure~\ref{visualization_st}(a), Principal Component Analysis (PCA) is used to compress the 250-dimension gene expressions into one dimension for better color mapping and visualization. As it is shown, slide-level methods such as HisToGene and Hist2ST tend to generate smoother ST maps, owing to the holistic processing of entire slides. In contrast, patch-level methods typically yield sharper predictions due to the independent processing of each spot in the ST map. Notably, M2OST consistently produces more accurate ST maps with distributions closely resembling the ground truth. This observation underscores the effectiveness of M2OST. Additionally, we augment our findings with individual gene visualizations in Figure~\ref{visualization_st}(b) to further elucidate the efficacy of M2OST. The gene we selected for visualization is DDX5, which plays a pivotal role in the proliferation and tumorigenesis of non-small-cell cancer cells by activating the beta-catenin signaling pathway \cite{wang2015ddx}. Our results indicate that M2OST achieves the highest accuracy in gene expression prediction for the selected gene, surpassing the performance of other patch-level and slide-level methods. 

% More visualization results can be found in Appendix-\ref{appendix-c}.

\section{Conclusion}

In this study, we tackle the challenging task of predicting ST gene expressions from WSIs by proposing a novel many-to-one-based regression Transformer, M2OST. M2OST leverages pathology images from several distinct levels to collectively predict gene expressions within their common central tissue area. The model incorporates M2OST Encoder for decoupled multi-scale feature extraction, which comprises ITMM for intra-scale representation learning, CTMM for cross-scale feature extraction, and CCMM for multi-scale channel mixing. The experimental results on three public ST datasets show that M2OST can achieve state-of-the-art performance with minimal parameters and FLOPs.

\section{Acknowledgments}
This work was supported by the National Key Research and Development Program of China (No. 2022YFC2504605). It was also supported in part by the Grant in Aid for Scientific Research from the Japanese Ministry for Education, Science, Culture and Sports (MEXT) under the Grant No. 20KK0234, 21H03470.
% \section*{References}

% \bibliographystyle{aaai25.bst}
\bibliography{aaai25.bib}
\newpage
\section{Appendix}
\appendix

\section{Selected Genes}

The names of the selected genes for each dataset are (Ensembl IDs have been converted into gene names for better readability) :

\noindent \textbf{HBC dataset:} PFKL, PLD3, ROMO1, PLXNB2, PCBP2, RPL19, HDLBP, RPL18, RPS4X, CD99, EIF5A, UBC, RPS3A, EEF1G, TLE5, RPS23, PSMD8, ACTN4, UQCR11, SRRM2, RPL8, BCAP31, GPX4, SCAND1, NDUFA3, TAPBP, RHOC, CYC1, COX5B, PTMA, GRN, RPL38, FTH1, TUBA1B, PABPC1, H2AJ, SPARC, NDUFA4, NDUFA11, UBA52, NENF, HLA-C, GPX1, EIF4A1, CLDN4, WDR83OS, SELENOW, MZT2B, RABAC1, RPS5, HM13, RPL24, HNRNPK, ENSA, GNAI2, SDF4, PPP1CA, JUP, HLA-A, FASN, RPS28, RPL27A, RPL37A, BEST1, EIF5AL1, TUFM, RPL11, RPL32, MYH9, RPS10-NUDT3, UQCRQ, TIMP1, NDUFB9, MYL6, RPS9, ATP1A1, EIF4G1, HLA-E, FLNB, FTL, NDUFB11, NDUFS6, RHOA, RPS10, KRT8, UBE2M, H1-10, C12orf57, ACTB, EDF1, PSMB4, ELOVL1, SPINT2, RPL14, ATP5F1E, RPS25, FBXW5, RPL13, PKM, EEF2, SLC2A4RG, ENO1, CD74, CENPB, TUBB4B, RPS13, CAPNS1, COLGALT1, SNRPD2, RPS18, RPL29, TMED9, CST3, RPL36A-HNRNPH2, CHD4, RPL31, RPS2, RPL30, RPL28, GAPDH, RPS19, LGALS1, RPL36A, GNB2, AP2S1, NDUFB7, SH3BGRL3, RPL36, PFN1, RPL12, MBOAT7, VCP, DDX5, ATP6V0C, MCL1, ADAMTSL4-AS1, RPLP1, EIF1, RPS12, EEF1D, RPL15, APOE, GNAS, SERF2, RPS16, RPL27, NDUFB10, DDX39B, ARHGDIA, LMAN2, SSR4, LAMP1, ADAR, COPS9, HNRNPA2B1, RPS15, RPL10, BRD2, ATP5F1B, CRIP2, LMNA, RPS29, AP2M1, CALR, ATP5MC2, SEC61A1, UBB, ATP6V1G2-DDX39B, CSDE1, COX6B1, MIF, LGALS3BP, PSMD2, CD81, COPE, PTPRF, ERGIC3, RPL18A, PEBP1, CHCHD2, RPL37, ZYX, EIF3B, SLC44A2, LSM4, NBEAL1, CTBP1, ELOB, FLNA, RPL7, PRRC2A, PRDX1, PFDN5, RPS17, RPS14, RPS24, CHCHD10, CTSD, MYO1C, RPL7A, KDELR2, RPSA, RPL35A, COX8A, RPS21, GRINA, CCT7, TPT1, RPL13A, NME2, RPL17, AUP1, PSAP, RPS6, HLA-B, ACTG1, MYDGF, EEF1A1, PHB2, RPL34, GUK1, RPL23A, FAU, RPS27, MALAT1, BSG, RPL22, ATP6V0B, RPS11, CHPF, RNASEK, FN1, RPL9, ATP6AP1, ZNF90, SLC25A6, RACK1, OAZ1, C18orf32, NCOR2, RPS27A, S100A6, CTSB, TMSB10, HSP90AB1, ALDOA, CFL1, RPS15A, COX4I1.

\begin{table*}[t]
\centering
\begin{tabular}{@{}c|cc|cc|cc|ccc@{}}
\toprule
Training Set & \multicolumn{2}{c|}{HBC}        & \multicolumn{2}{c|}{HER2+}     & \multicolumn{2}{c|}{cSCC} & \multicolumn{3}{c}{All} \\ \midrule
Testing Set  & HER2+          & cSCC           & HBC            & cSCC          & HBC             & HER2+   & HBC   & HER2+   & cSCC  \\ \midrule
ST-Net       &  25.31   &  6.57    &   25.61    & 12.55     & 6.22     &  13.19       &  34.59     &   35.08      & 33.24      \\
HIPT/iStar      & 26.46 & 6.95 & 26.78       & 13.01    & 6.72  &  16.44   &   37.92     & 38.29        &  36.10    \\
Hist2ST  & 21.04 & 3.59 & 23.75       & 10.72    & 3.17  &  12.64   &   33.56     & 33.17        &  32.70  \\
M2OST (Ours)       & \textbf{27.39}   &  \textbf{9.92}    &   \textbf{28.00}    &  \textbf{14.58}     &  \textbf{9.37}        &  \textbf{17.54}       &  \textbf{39.46}     & \textbf{39.97}        & \textbf{37.33}     \\ \bottomrule
\end{tabular}
\caption{PCC results of external validation experiments.}
\label{external_validation}
\end{table*}

\noindent \textbf{HER2+ dataset:} HLA-C, DDX5, TSPO, PRPF19, HLA-DRA, LMNA, APOC1, HDLBP, PSMB4, CIB1, NECTIN2, C19orf53, ENSA, CLDN7, CST3, PRDX6, ATP5G2, CDC37, UBC, SLC44A2, GIPC1, MAPKAPK2, PFDN5, TCEB2, COL18A1, CLDN3, SNRPB, H1FX, MZT2B, TMED9, CLIC1, UBA52, SCAND1, LAMTOR4, FKBP2, SEPW1, TRAF4, NDUFB2, H3F3B, S100A6, NDUFB11, TUFM, FXYD3, SSR4, NRBP1, YIPF2, PPP2R1A, GRB7, ATP5E, PSMB7, SLC9A3R1, WDR34, BSG, ERBB2, VIM, NUMA1, MAP2K2, COPS9, MAGED1, PHB, IDH2, CRIP2, COPE, AUP1, NUPR1, ELOVL1, S100A11, COX4I1, ACTG1, XBP1, MMACHC, COL1A1, AP2S1, HNRNPA2B1, STUB1, TAGLN, NDUFS6, MDK, UBB, OAZ1, BEST1, KDELR2, EDF1, CD24, CCND1, ATOX1, RER1, HSP90B1, CTSB, SNRNP200, C14orf2, FTL, PPP1CA, CCT7, P4HB, HSBP1, SDF4, SNRPD2, COX8A, TAPBP, SNF8, APOE, PTMA, CYBA, VCP, ECHS1, ARPC1B, ROMO1, ZYX, UQCRQ, ACTB, CYC1, PRRC2A, INF2, MIDN, SPINT2, TPI1, PTBP1, NDUFA4, PRSS8, PTPRF, GNAS, PKM, PLD3, SLC39A1, HSPG2, MGP, RALY, COX7C, LAPTM5, MT2A, CCT3, NENF, UQCR11, TIMP1, ATP6V0B, ARPC4, NME3, PRKCSH, ARRDC1, KRT8, CTBP1, RHOA, MYO1C, HDGF, GPX4, FTH1, ATP5J2, JUNB, PEBP1, GAPDH, GLTSCR2, BRD2, KRT7, TMEM219, HLA-A, FN1, EIF3K, FBXW5, LGALS1, CTSD, ENO1, NDUFB9, KDELR1, SLC2A4RG, RABAC1, EIF3B, JTB, C12orf57, STARD3, HSP90AA1, ATP5I, GNAI2, AES, UBL5, LMAN2, MCL1, AEBP1, EEF1D, GUK1, FAU, ATP5J, PSMD2, CISD3, HNRNPAB, BGN, TUBA1B, HLA-B, PFN1, CHCHD2, SEPT9, NR2F6, ZBTB7B, PDXK, MGAT1, ATP5B, TPT1, EIF4G2, DNAJB1, CD81, NDUFA10, PPP1R14B, RHOC, OST4, SLC25A6, HSP90AB1, MRPS34, PRMT1, CD63, EIF4G1, CAPZB, CALR, MZT2A, TUBB, CFL1, NPDC1, SRRM2, COL3A1, PFKL, AKT1, RACK1, RRBP1, APRT, CALM3, DDB1, KRT19, CTTN, CD99, HES4, CENPB, MUC1, SEC61A1, ATP1A1, COX6B1, PTOV1, UQCR10, CERS4, EEF2, SERF2, GRINA, BCAP31, LSM4, NDUFS5, LSM7, KRT18, TIMM13, MFSD12, SH3BGRL3, SHC1, ARHGDIA.

\noindent \textbf{cSCC dataset:} VAPA, PPP4R1, TGM1, GNAI2, CD82, MYO1C, DBI, PSMA4, PPP1CA, HSP90AA1, COX6A1, BICD2, PSMB4, ITGA3, RPL32, RPS18, NECTIN1, MAPK6, SYNCRIP, FGFBP1, CFL1, CD81, DAZAP1, PDIA3, TXNIP, CSNK2B, ACTN4, HNRNPK, RPL13, TPI1, SURF4, DSG3, ARPC1A, RAB11A, RPS12, STAT1, TOMM7, ADGRG1, LGALS7, TRIP12, LAPTM4A, RPL6, PSMD2, RPL18, MCL1, MAP2K2, ANXA5, APRT, C4orf3, MALL, SLC2A4RG, TYMP, PDIA6, POLR1D, AP2M1, CCND1, ARHGEF12, NPEPPS, H3F3B, S100A16, PLP2, PTRF, SNRPD2, KLF5, HNRNPD, CYFIP1, RNH1, NCL, CNFN, TAPBP, TCEB2, DHCR24, EIF3CL, ZFP36L2, FSCN1, CHD4, KLF6, MLF2, FOSL2, MT-CO3, RPL26, HLA-C, PSMB3, NOMO2, ATP5O, CAST, FXYD3, NDUFB4, MSMO1, HMGA1, RPL19, ANXA1, SPRR1B, C6orf62, KIF5B, ARPC2, NME1, PSMD1, RPS23, CEBPB, MAF, RPS6, CD63, UBE2D3, RPS19, MT2A, PGK1, MT-CO2, PRRC2C, NHP2, TXN, LARP1, TMEM165, GPNMB, SERP1, S100A8, DMKN, GSN, ETF1, GNAS, CDSN, CXCL14, EIF5A, ROMO1, RPL7A, KRT6C, DSTN, LRP10, EIF3B, VAMP8, NDUFB10, NACA, IER2, SRSF5, MARCKS, REEP5, SAP18, MT-ND2, UQCRC1, KLK11, EDF1, YWHAG, CALM1, KRT17, RPSA, MT-CO1, RPL29, CDC37, COL18A1, HDLBP, KRT6A, PTGES3, RPS2, ATP5I, ILF3, PSAP, JAG1, LAMP1, RPL36, RPS27A, GNB1, NPM1, RPLP2, NDUFS5, NDUFB1, FLNB, PSMC3, TMED10, MIDN, VCP, CSTB, QARS, FABP5, CAPN1, RPS8, POLR2L, TAGLN2, COX8A, HLA-E, PFDN5, HMGB1, ARF6, DSC3, DSC2, ZFP36L1, EIF4G1, CD9, FTL, COL7A1, H1FX, SLC7A5, KDELR1, RPS3A, SEPW1, TPSB2, RPL28, KHDRBS1, ACTR2, APP, PTMA, SON, PPL, SERF2, GRN, VIM, MAFB, CRCT1, KTN1, PSMB6, STUB1, RPS14, LAMP2, VMP1, BCAP31, IFI16, PTP4A1, TMA7, ATP5J2, CANX, SRSF11, ITGAV, FLNA, ITGA6, COX7C, DST, MAL2, ACTN1, RPL9, RPL36A, ZNF207, TMED9, RNF187, PSMA7, EIF3H, RPS25, WNK1, MYL12A, RPS26, FUS, LCE3D, CNDP2, MYL12B, IVL, IGFBP4, RPS5, RPL11, MKNK2, RACK1, FTH1, EEF1B2.

In practical applications, doctors have the flexibility to manually specify the target genes before the training process. This enables predictions tailored to their specific analysis needs, enhancing the utility of the model in real-life scenarios.

For external validation, we use the intersection of the 250 genes in the three datasets, which leads to a total number of 33 genes.

\begin{table*}[t]
\renewcommand\tabcolsep{6pt}
\centering
\begin{tabular}{@{}ccccccccc@{}}
\toprule
\multirow{2}{*}{\begin{tabular}[c]{@{}c@{}}Masked \\Percentage\end{tabular}}  & \multicolumn{2}{c}{HBC} & \multicolumn{2}{c}{HER2+} & \multicolumn{2}{c}{cSCC}  & \multirow{2}{*}{\begin{tabular}[c]{@{}c@{}}Mean \\PCC (\%)\end{tabular}} & \multirow{2}{*}{\begin{tabular}[c]{@{}c@{}}Mean \\RMSE\end{tabular}} \\ \cmidrule(lr){2-3} \cmidrule(lr){4-5} \cmidrule(lr){6-7}
                         & PCC(\%)      & RMSE     & PCC(\%)       & RMSE      & PCC(\%)      & RMSE   \\ \midrule
0\%            & \textbf{39.46}      & \textbf{4.20}        &   \textbf{39.97}            & \textbf{4.45}          & \textbf{37.33}       &  \textbf{4.01}  & \textbf{39.27} & \textbf{4.19} \\
25\%            & 37.15    & 5.25    &  36.26          & 5.62        & 36.01       & 4.76  & 36.51 & 5.30 \\
50\%              & 36.54     & 5.65        &  32.42            & 6.02          & 33.54       &  5.00 & 35.01 & 5.62  \\
75\%            &   31.71    & 5.68       &     32.32        & 6.01       &  34.19   & 5.00 & 32.32 & 5.63  \\ \bottomrule
\end{tabular}
\caption{Performance of M2OST after masking one random level of image in different percentages of the samples in the fused dataset (i.e., the dataset combining HBC, HER2+ and cSCC together). }
\label{random_drop_result} 
\end{table*}

\begin{table*}[t]
\renewcommand\tabcolsep{3.2pt}
\centering
\begin{tabular}{@{}ccccccc@{}}
\toprule
\multirow{2}{*}{Backbone} & \multicolumn{2}{c}{HBC} & \multicolumn{2}{c}{HER2+} & \multicolumn{2}{c}{cSCC} \\ \cmidrule(l){2-3} \cmidrule(l){4-5} \cmidrule(l){6-7} 
                          & PCC(\%)      & RMSE       & PCC(\%)      & RMSE        & PCC(\%)      & RMSE    \\ \midrule
ResNet-50   & 48.81 (+1.71)       &  3.16 (-0.01)       &   44.76 (+1.43)            &  2.86 (-0.18)         &   50.80 (+1.46)           &  3.41 (-0.19)      \\
ST-Net   &    48.90 (+1.12) & 3.16 (-0.00)   & 44.73 (+1.72) & 2.90 (-0.17)          &  50.76 (+1.39)   &  3.43 (-0.15) \\
ViT           & 48.62 (+1.95) &  3.17 (-0.00)  & 44.41 (+0.63)  &  2.97 (-0.12) &  50.82 (+1.81)   & 3.44 (-0.33)  \\ \bottomrule
\end{tabular}
\caption{Experimental results of adopting many-to-one modeling scheme to other backbones.}
\label{m2o_on_other_backbones}
\end{table*}

\section{Additional Experimental Results}

% \subsubsection{Study on the Scalability of M2OST Structure.} New table needed.

\subsection{External Validation}

External validation is crucial for the practical application of M2OST. Without employing any domain generalization or domain adaptation techniques, we have performed a thorough external validation for M2OST and other methods. In the experiment, one entire ST dataset is used for training, while the rest two datasets are used for validation. The experimental results are presented in Table~\ref{external_validation}. For a more comprehensive comparison, we have also randomly selected 70\% cases in each dataset for joint training, and have presented the testing results on each dataset in the table as well.

The results indicate that transferring from cSCC to HBC/HER2+ or from HBC/HER2+ to cSCC is generally more challenging than other cases. This is because cSCC is for cutaneous squamous cell carcinoma while HBC and HER2+ datasets are both in the breast cancer domain. In general, M2OST shows better generalization ability than other methods. Furthermore, it also achieved the best performance when trained on all three datasets, illustrating the effectiveness of the many-to-one multi-scale modeling design.

\subsection{Dealing with Missing Levels in M2OST}

As aforementioned, M2OST differs from conventional multi-scale methods in its ability to be flexibly scaled to accommodate various many-to-one scenarios. To validate M2OSTs effectiveness under conditions where certain levels of input images are missing, we have conducted an experiment. In this experiment, we randomly removed one level of the images from 25\%, 50\%, and 75\% of the samples in the fused dataset (i.e., the dataset used in Table~\ref{external_validation}, which combines all three datasets). The missing images were replaced with all-black images and were detached during loss back-propagation. The predicted output, $\hat{G}$, is obtained by averaging the prediction results from each level. During inference, samples with missing image levels produce the final prediction $\hat{G}$ based on the mean value of the gene expression predictions from the remaining two levels.

The experimental results are presented in Table~\ref{random_drop_result}. As shown, M2OST effectively models the many-to-one relationship even when some inputs are missing, demonstrating the flexibility of its structure. Specifically, we observed that as the number of masked samples increases, M2OST experiences greater performance loss. When random masking was applied to 75\% of the samples, the models PCC dropped by 7.75\%, 7.65\%, and 3.14\% on the three datasets, respectively. This performance decline is attributed to the significant information loss, making it challenging for M2OST to accurately predict gene expressions.

\subsection{Study on Adopting Many-to-one Scheme to Other Backbones}

From the experimental results in Table 2 and Table 3 of the main paper, it is observed that M2OST cannot yet surpass ordinary backbones when being used as a one-to-one method. However, after introducing the proposed many-to-one scheme to it, M2OST achieved state-of-the-art performance with minimal computational cost. This demonstrates the effectiveness of our proposed many-to-one modeling scheme, and raises the question of whether this approach can be applied to other existing backbones for further performance improvements.

However, unlike M2OST being designed especially for many-to-one modeling, existing backbones typically cannot process multiple inputs simultaneously, complicating the adoption of the many-to-one modeling scheme. To address this, we used the representations extracted by these existing backbones as inputs for M2OST. Specifically, the three levels of pathology patches were directly input into the existing backbones for patch embedding, and the extracted representations were then used as multi-level sequences for M2OST to perform many-to-one modeling.

The experimental results are presented in Table~\ref{m2o_on_other_backbones}. As shown, using these backbones for patch embedding leads to further performance improvement compared with the original M2OST. On one hand, this indicates that these existing backbones are able to generate more accurate token representations for M2OST, and on the other hand, it also demonstrates that many-to-one modeling can further improve the performance of these backbones. In our future works, we will investigate how to more efficiently introduce the many-to-one scheme to existing backbones.

% \section{Additional Visualization Results}
% \label{appendix-c}
% To further illustrate the effectiveness of M2OST, we have presented some additional visualization results in Fig~\ref{per_gene_visualization_sup} and Fig~\ref{visualization_st_sup}. These results further demonstrate the effectiveness of M2OST for the ST prediction task.

\section{Limitations} 
As a spot-level method, M2OST has a lower efficiency compared to the slide-level methods, as it can only generate gene predictions of one spot at a time. Additionally, since M2OST is an end-to-end spot-level method, it can only perceive limited nearby non-local information in a WSI during training, which may leave space for future improvements.

\end{document}